\documentclass[runningheads]{llncs}
\usepackage{graphicx}
\usepackage{amsmath}
\usepackage{hyperref}
\usepackage{appendix}
\usepackage{multirow}
\usepackage{tabularx}
\usepackage{makecell}
\usepackage{cite}
\usepackage{stfloats}
\usepackage{longtable}
\usepackage{lipsum}
\usepackage{float}
\usepackage{afterpage}
\usepackage{amsmath,amssymb,amsfonts}
\usepackage{algorithmic}
\usepackage{multirow}
\usepackage{hyperref}
\usepackage{caption}
\usepackage{graphicx}
\usepackage{textcomp}
\usepackage{xcolor}
\usepackage{lipsum}
\usepackage{booktabs}
\usepackage{multirow}
\usepackage{algorithm}
\usepackage{algorithmic}
\usepackage{xcolor}
\usepackage{marvosym}
\hypersetup{
    colorlinks=true,
    linkcolor=red,
    urlcolor=blue,
    citecolor=green
}

\begin{document}

\title{
    FGeo-TP: A Language Model-Enhanced Solver for Geometry Problems
}
\titlerunning{FGeo-TP: A Language Model-Enhanced Solver for Geometry Problems}

\author{
    Yiming He\inst{1,2} \and
    Jia Zou\inst{1,2} \and 
    Xiaokai Zhang\inst{1} \and
    Na Zhu\inst{1,2} \and
    Tuo Leng\inst{1,2}\textsuperscript{(\Letter)}
}
\authorrunning{X. He et al.}

\institute{
    School of Computer Engineering and Science, Shanghai University, Shanghai, China \\ \email{tleng@shu.edu.cn} \and
    Institute of Artificial Intelligence, Shanghai University, Shanghai, China 
}

\maketitle

\begin{abstract}
The application of contemporary artificial intelligence techniques to address geometric problems and automated deductive proof has always been a grand challenge to the interdiscipline field of mathematics and artificial Intelligence. This is the fourth article in a series of our works, in our previous work, we established of a geometric formalized system known as FormalGeo. Moreover we annotated approximately 7000 geometric problems, forming  the FormalGeo7k dataset. Despite the FGPS (Formal Geometry Problem Solver) can achieve interpretable algebraic equation solving and human-like deductive reasoning, it often experiences timeouts due to the complexity of the search strategy. In this paper, we introduced FGeo-TP (Theorem Predictor), which utilizes the language model to predict theorem sequences for solving geometry problems. We compared the effectiveness of various Transformer architectures, such as BART or T5, in theorem prediction, implementing pruning in the search process of FGPS, thereby improving its performance in solving geometry problems. Our results demonstrate a significant increase in the problem-solving rate of the language model-enhanced FGeo-TP on the FormalGeo7k dataset, rising from 39.7\% to 80.86\%. Furthermore, FGeo-TP exhibits notable reductions in solving time and search steps across problems of varying difficulty levels.


\end{abstract}

\section{Introduction}

The utilization of computers to solve mathematical problems has long been an intriguing and highly challenging endeavor. With the continuous evolution of artificial intelligence technology, various methods have emerged for solving mathematical problems across different domains. This has led to the creation of solvers specifically designed for arithmetic, algebraic ~\cite{drori2022neural}~\cite{mundhenk2021symbolic}~\cite{fawzi2022discovering}~\cite{polu2022formal} , and theorem proving ~\cite{yang2023leandojo}~\cite{polu2020generative}. While there has been relatively limited research on plane geometry problems, recent years have witnessed the emergence of corresponding studies ,such as inter-GPS~\cite{lu2021intergps},GeoQA~\cite{chen2021geoqa},uniGEO~\cite{Chen2021UniGeo}.In our previous research, FormalGeo~\cite{Zhang2023FormalGeo}, we outlined how artificial intelligence could be utilized for automated reasoning in solving plane geometry problems. Previous efforts involved the construction of the FormalGeo7k dataset and the development of FGPS (Formal Geometry Problem Solver).

\begin{figure}[t]
  \centering
  \includegraphics[width=0.8\textwidth]{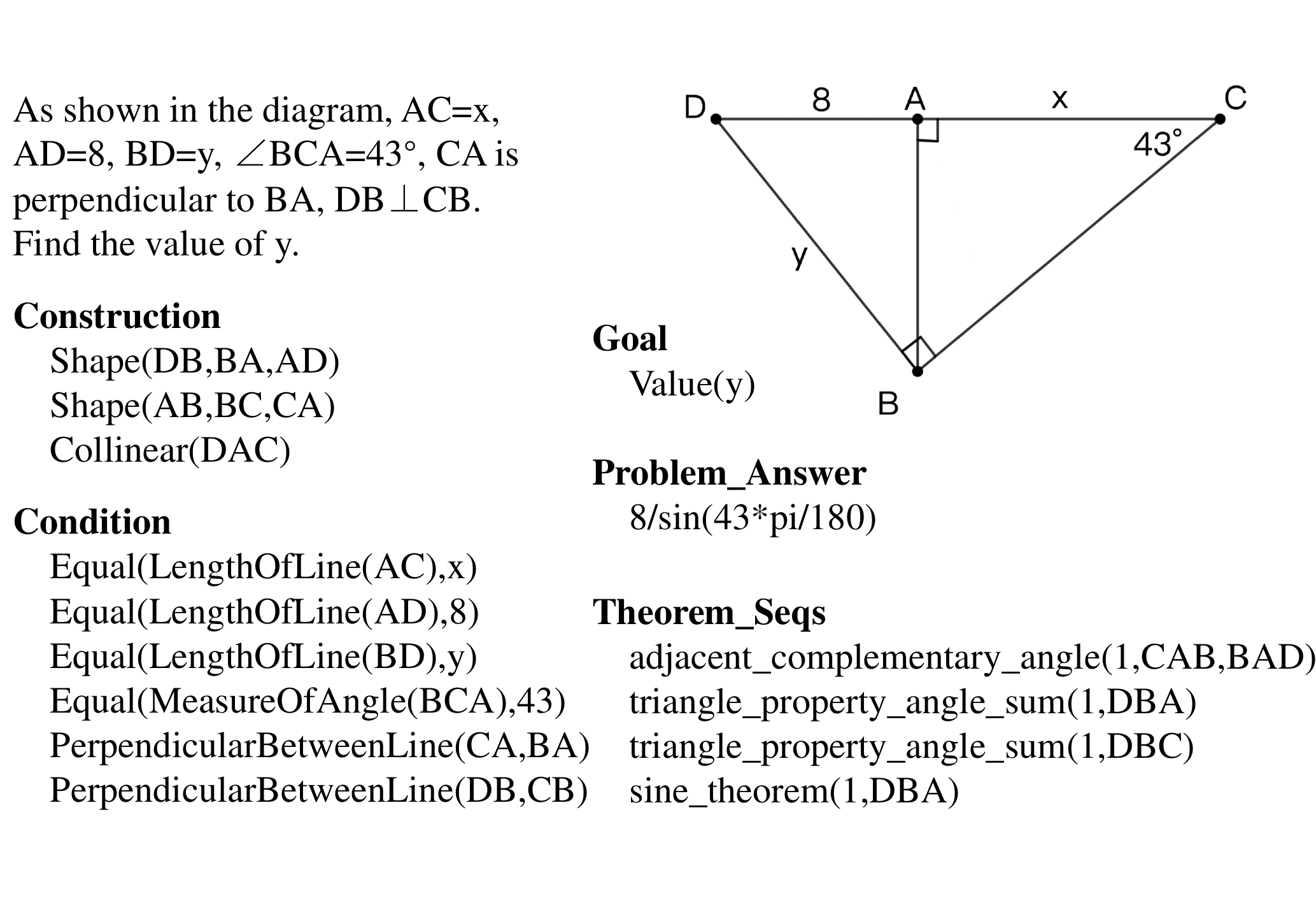}
  \caption{An example from FormalGeo7k, including formal annotations.}
  \label{fig:fig1}
\end{figure}

The construction of the FormalGeo7k dataset was grounded in geometric ontology and geometric representation theory. It comprises 6,981 SAT-level geometry problems, each accompanied by a complete natural language description, geometric shapes, formal language annotations, and theorem sequence annotations. One of the examples is illustrated in Figure \ref{fig:fig1}.When provided with the annotated theorem sequences, FGPS is capable of solving all problems in the FormalGeo7k dataset and returning the problem-solving process. Therefore, all subsequent mentions of FGPS's problem-solving in this article refer to results obtained without the provided annotated theorem sequences, relying solely on input formalized conditions.

The FGPS  employs both forward and backward search methods with various strategies for automated problem-solving. It is capable of executing traceable and interpretable algebraic equation solving and relationship reasoning. Experiments conducted on the FormalGeo7k dataset indicate that FGPS achieved a maximum solving rate of 39.7\% with the forward random search method.Both forward and backward search methods showed success rates of less than 40\% in solving correct answers within a limited time frame. To achieve true artificial intelligence automated reasoning in solving plane geometry problems, auxiliary measures are needed to enhance the speed and effectiveness of FGPS.

Inspired by existing works , we found that language models can effectively comprehend the annotations of geometric information in the FormalGeo7k dataset. In the dataset, the formalized representation of geometric conditions takes the form of statements such as "Equal(LengthOfLine(AD),8)," indicating the length of line segment AD is 8. Our formalization is easily comprehensible by both humans and computers. Therefore, we propose integrating FGPS with language models, leveraging the inferential capabilities of language models to train them in predicting theorem sequences required for geometric problem-solving. This augmentation aims to enhance the solving efficiency of FGPS. We employ various language models for prediction and select the most effective result among them as the predicted theorem sequence to be incorporated into the FGPS solving process. The specific approach involves predicting the theorems that may be required for a given geometric problem before FGPS initiates its theorem search. The solver can execute the predicted theorems first, incorporating the conclusions obtained through theorem execution into the set of conditions. This ensures a reduction in the steps and time required for the solver to utilize forward or backward search strategies.

In summary, our contributions are three-fold:

1.We propose the combination of FGPS and language models to streamline the search process.

2.Regarding the FormalGeo7k dataset, we employed multiple language models for theorem sequence prediction, achieving a high degree of theorem sequence matching.

3.Implemented FGeo-TP (FGPS and Theorem Predictor), significantly improving the success rate of problem-solving compared to FGPS and greatly reducing search time and steps.

\section{Related Work}

\subsection{Datasets for Geometry Problem Solving}

In recent years, numerous exemplary planar geometry datasets have emerged. However, we contend that the formalization methods employed in these datasets lack uniformity. This includes datasets such as Geometry3K ~\cite{lu2021intergps}, GeoQA, and PGDP5K ~\cite{hao2022pgdp5k}. Each of these datasets employs distinct formalization methods for annotating geometric problem-solving. Geometry3K annotates information in problem statements using both diagram formal language and text formal language but does not provide annotations for solution information. PGDP5K is designed to construct a formal language dataset through geometric image analysis. However, these approaches lack concrete mathematical theoretical support, leading to a deficiency in ensuring completeness.The GeoQA dataset encompasses nearly 5000 planar geometry problems, predominantly focused on numerical calculations involving angles and lengths. However, it lacks data related to geometric relationship proofs. Moreover, answer annotations in these datasets often manifest as multiple-choice questions, introducing the possibility of solvers randomly selecting the correct answer.

In contrast, our FormalGeo7k dataset consists of 6,981 SAT-level geometry problems, further expanded to 186,832 through data augmentation. Each problem in the dataset includes a comprehensive natural language description, geometric shapes, formal language annotations, and theorem sequence annotations. Importantly, our dataset does not adopt a multiple-choice question format but instead presents authentic planar geometry problems akin to those encountered by secondary school students in routine assessments. The answer types encompass numerical values, geometric relationships, and combinations thereof. 

The FormalGeo7k dataset is aggregated from diverse sources, including Geometry3K, GeoQA, GeoQA+~\cite{cao2022augmented}, and online repositories. We meticulously curated, classified, deduplicated, and standardized the problem statements. Creating the FormalGeo7k dataset was a substantial undertaking, which involved approximately 16 trained master's students over a period of around 13 weeks. Excluding collaboration and dataset allocation time, the annotation of problem statements required approximately 700 person-hours each.

Furthermore, the construction of the FormalGeo7k dataset is guided by studies in Geometry Ontology and Geometry Representation Theory. These methodologies address the questions of what content should be formalized and how it should be formalized. Consequently, the FormalGeo7k dataset is both extensible and readily interpretable, distinguishing it from other geometry datasets mentioned earlier, which often lack scalability.

\subsection{Geometry Problem Solving}

The history of computer-aided geometric problem-solving can be traced back to the previous century. Gelernter et al. employed a backward search method to address formalized problems\cite{gelernter1963realization}, and Nevins utilized the forward chaining method\cite{nevins1975plane}. Search-based methods often prove only a limited number of planar geometry theorems due to their high computational complexity. Wen-Tsun proposed Wu's Method\cite{wen1978decision}, which transforms geometry problems into algebraic equation-solving problems but is confined to the algebraic domain. Zhang introduced the point elimination method based on geometric invariants \cite{zhang1995automated}, generating concise and meaningful readable proofs for a large number of geometry problems. However, this method is restricted by the types of geometric invariant points.

In recent years, Seo et al. established the first automated system, Geos\cite{seo2015solving}, which translates the text and images of geometry problems into logical language. However, its high dependence on manually annotated rules limits its generalization. GeoQA transforms the process of solving geometry problems into a sequence of programs composed of variables and operators, resulting in problems with lower readability compared to formal languages. Lu et al.  proposed Inter-GPS, achieving high accuracy across multiple geometry datasets. However, Inter-GPS is limited to numerical calculation problems and cannot handle geometry relationship proofs.

In our prior work, FGPS can reason through all geometry problems in the FormalGeo7k dataset, providing detailed solution processes. Specifically, for a given planar geometry problem, FGPS can output a complete solution process, including the mathematical theorems used in each step, which conditions of the problem are involved in theorem application, and the new geometric conditions generated after applying the theorems. FGPS can emulate the step-by-step problem-solving process as a secondary school student approaching planar geometry problems. Moreover, FGPS is not limited to numerical calculation problems and can effectively handle geometry relationship proof problems.

\subsection{Pre-training Model In NLP}

In the field of Natural Language Processing (NLP), pre-trained models play a crucial role by enabling NLP models to achieve excellent performance without the need to train from scratch. Instead, these models can undergo fine-tuning directly on pre-trained counterparts, streamlining the training process and yielding outstanding results\cite{wu2020knowledge}\cite{sun2020colake}.
In the work of MWp-bert\cite{liang2021mwpbert}, pre-trained models were employed for solving mathematical word problems. GeoQA+ utilized pre-trained models to transform the textual information of geometric problems, thereby augmenting the dataset. Inter-GPS used pre-trained models to predict theorems; however, the Geometry3K dataset in Inter-GPS did not have manually annotated theorem sequences required for solving problems. Instead, they were randomly sampled, leading to potentially non-optimal theorem sequences. Furthermore, it could only predict theorems for 1500 questions and not for all questions in the dataset.

The Transformer\cite{vaswani2017attention} is a neural network model used for sequence-to-sequence\cite{sutskever2014sequence} learning, which classic Encoder+Decoder architecture has shown excellent results in NLP tasks. Inspired by this success, we propose using the Transformer architecture to comprehend formal statements and predict the theorems required for geometric problem solvers.

The formalized conditional statements in FormalGeo7k's geometry problems are easily understandable, not completely detached from the computer's comprehension of English text. Since Transformer demonstrates remarkable effectiveness in understanding English text, we consider employing the Encoder+Decoder architecture of Transformer to learn the relationship between conditional formalized statements and theorems. For the seq2seq task, we conducted a comparative analysis of various language models to determine their performance.

\section{Geometry Problem Solver}
\subsection{FGPS}
In our previous work\cite{Zhang2023FormalGeo}, we implemented FGPS for inference and solution of all problems in the FormalGeo7k dataset. However, this was contingent on the annotated theorem sequences provided in the FormalGeo7k dataset. When the solution theorem sequence is not provided, FGPS resorts to using built-in methods to search for theorem solutions. During the search process, FGPS constructs a search tree containing the sequence of theorem applications for solving a given problem. We utilized both forward search (FW) and backward search (BW) methods. FW starts from known conditions, continually searching for available theorems to generate new conditions until the solving objective is reached. In contrast, BW starts from the solving objective, decomposes it into multiple sub-goals, and seeks the conditions required for each sub-goal, determining whether the current sub-goal is solvable. This process repeats until all sub-goals are resolved.

Additionally, we employed four search strategies: Breadth-First Search (BFS), Depth-First Search (DFS), Random Search (RS), and Beam Search (BS). BFS traverses each node of the search tree in a level-wise manner, DFS recursively selects nodes from shallow to deep, RS randomly selects nodes for expansion at each stage, and BS selects a specified number of nodes (K) at each expansion stage, striking a balance between BFS and RS.

In cases where the search time exceeded 600 seconds, indicating a timeout for problem-solving, the search results are presented in Table 1. Notably, the forward random search method achieved the highest success rate of 39.7\%, while the backward depth-first search method exhibited the lowest unsolved rate of 2.42\%. We observed that a substantial portion of problem-solving tasks were not entirely unsolvable but rather failed due to prolonged solving times, leading to timeout. Hence, there is a need for optimizations and pruning of the FGPS solving process to achieve a higher success rate in problem-solving.

\begin{table}[h]
  \centering
  
  \begin{tabular}{c|c|c|c|c}
    \hline
    method & strategy & solved & unsolved & timeout \\
    \hline
    FW & BFS & 38.86 & 7.42 & 53.72 \\
    FW & DFS & 36.16 & 9.80 & 54.05 \\
    FW & RS & \textbf{39.71} & 9.07 & 51.22 \\
    FW & BS & 25.28 & 38.72 & \textbf{36.00} \\
    BW & BFS & 35.44 & 2.68 & 61.88 \\
    BW & DFS & 33.73 & \textbf{2.42} & 63.84 \\
    BW & RS & 34.05 & 2.65 & 63.30 \\
    BW & BS & 34.39 & 12.86 & 52.74 \\
    \hline
  \end{tabular}
  \caption{The search results(\%) of FGPS}
  \label{tab:result}
\end{table}

 In line with the habitual problem-solving practices of humans, a high school student, accustomed to regular problem-solving, typically skims through the problem conditions when faced with a plane geometry question. With this initial scan, the student can often make an approximate inference regarding the primary knowledge points being tested by the question. Therefore, our aim is for FGPS to emulate this cognitive process. To achieve this, we have incorporated a theorem predictor ahead of the solver in our methodology. This modification enables FGPS to select more suitable theorems for application, rather than attempting to utilize all available theorems.
\subsection{FGeo-TP}

\begin{figure}[h]
  \centering
  \includegraphics[width=0.8\textwidth]{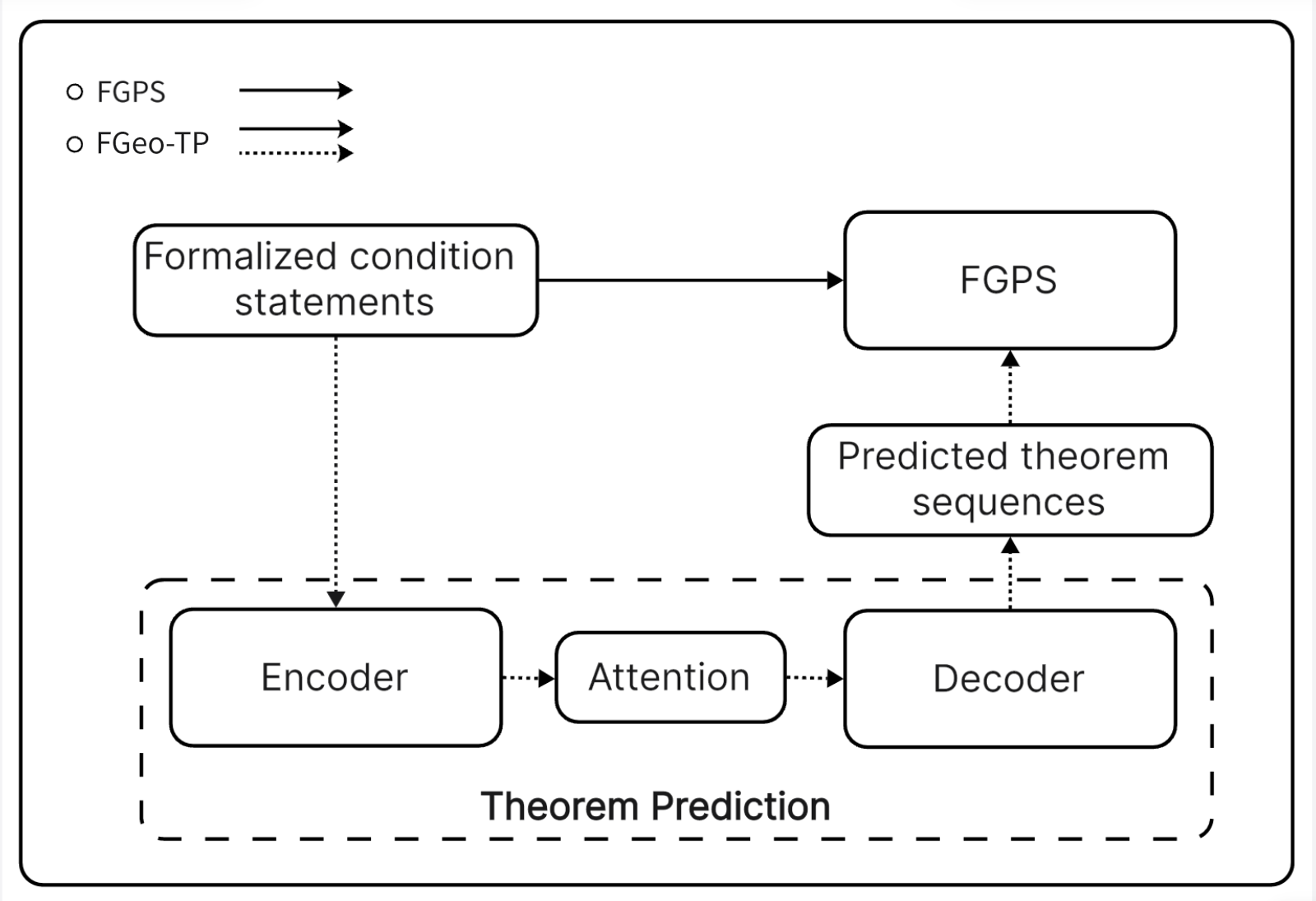}
  \caption{The architecture of FGeo-TP}
  \label{fig:fig2}
\end{figure}

FGPS experienced a high percentage of timeouts in searching the FormalGeo7k dataset, primarily because, during the search process, each step involves exploring a large number of theorems for potential matches. To optimize the solver's solving process, we introduced a theorem predictor. Before FGPS initiates theorem search, the theorem predictor guides FGPS, reducing the search complexity. The augmented FGPS, incorporating the theorem predictor, is denoted as FGeo-TP (Theorem Predictor). The architecture of FGeo-TP is illustrated in Figure ~\ref{fig:fig2}.

In contrast to directly inputting the formalized language into FGPS, FGeo-TP requires the formalized language to be simultaneously input into the theorem predictor. The theorem predictor outputs the corresponding theorem sequence, and FGPS receives both the formalized language and the predicted theorem sequence.

In the theorem predictor, we anticipate the input and output to be the formalized language from FormalGeo7k and the required theorem sequence for each problem. As the length of the formalized language and the theorems varies for each problem, a Seq2Seq model is considered suitable. We opted to use the well-established Transformer architecture for implementing both the Encoder and Decoder.

The formalized conditional language in FormalGeo7k consists of geometric relationship predicates, geometric form predicates, free variables, or numbers. This differs significantly from popular natural language corpora in the field of natural language processing. Therefore, pre-training the Transformer model with our corpus is essential for subsequent comprehension of formalized language by the model. To enhance readability, the formalized geometric relationship and form predicates in FormalGeo7k are composed of English words or concatenations of English words. This can be treated as word-level tokenization in the Encoder, while free variables and numbers are considered character-level tokenization. Hence, there is no need to update the existing vocabulary. Due to the specificity of theorem names, a slight change in a single word may render FGPS unable to recognize them. To prevent out-of-vocabulary situations in the Decoder, we convert theorem names into their corresponding numerical representations and represent the theorem sequence as a one-dimensional array containing only integers.

We performed fine-tuning on the Transformer pre-trained model using annotated training and validation datasets. The fine-tuned model was then evaluated on the test set to assess the prediction results. In fine-tuning tasks, the optimization was carried out using the Negative Log-Likelihood Loss to refine the generated objectives.

\begin{equation}
\text{Loss} = -\frac{1}{N} \sum_{i=1}^{N} \log P(y_i | y_1, \ldots, y_{i-1}, \text{input})
\end{equation}

N represents the length of the sequence, representing the number of theorems in the target sequence.The input represents The input formalized language.
To achieve higher matching degrees, we employed beam search, selecting the union of multiple theorem sequences with higher probabilities to enhance the degree of matching.

\begin{equation}
\quad B_{n} = \underset{y_1, \ldots, y_{n-1}, y_n}{\operatorname{argtopk}} \left[ P(y_n | y_1, \ldots, y_{n-1}, \text{input}) \right]
\end{equation}
 
\( B_n \) represents the set of the top k predicted theorem sequences when predicting the next theorem. Each predicted theorem sequence has the form [\( y_1, \ldots, y_{n-1},y_{n}\)], where   \( y_n \) is the theorem added in that step. The input is  the formalized geometry problem conditions. This process continues until the predefined sequence length is reached or an end token is encountered. In the experiments, we set the maximum sequence length to 20, and the number of top predicted theorem sequences is 5.

\begin{equation}
 S_{\text{tp}} = B_n{}_1 \cup B_n{}_2 \cup \ldots \cup B_n{}_k
\end{equation}

\(Stp\) represents the final predicted theorem sequence, while \(B_{n_1}, B_{n_2}, \ldots, B_{n_k}\) denotes the set of the top-k predicted theorem sequences when the last theorem prediction is completed. Experimental findings indicate that even when merging multiple predicted theorem sequences, the final predicted theorem sequence's length does not experience a significant increase.

After FGeo-TP executes the predicted theorem sequence, conclusions are deduced based on the original problem conditions. We integrate these conclusions with the initial problem conditions to form a new set of conditions. If the problem-solving objective is still unresolved with this updated set, FGeo-TP employs a search method based on the new condition set to identify a solution. FGPS demonstrates a quick use of the theorem, thus, the use of predicted theorem sequences by FGeo-TP does not impose a significant burden on reasoning. On the contrary, it can reduce FGPS's search for subsequent theorems. Through these steps, our objective is to streamline the solver's search process and implement pruning in the solver's solving procedure.

\section{Experiment} 
\subsection{Theorem Sequence Prediction}
We utilized data from the FormalGeo7k dataset for training purposes. Initially, we randomly shuffled the 6,981 geometry problems, allocating them to training, validation, and test sets in a ratio of 0.7:0.15:0.15. For the theorem prediction model, we attempted to compare various pre-trained language models based on the Transformer architecture.

It's worth noting that our initial experiments did not directly assess the problem-solving effectiveness of FGeo-TP. Instead, we began by evaluating the accuracy of predicting theorems for various pre-trained language models. We trained the models using formal language annotations and theorem sequences annotations from the FormalGeo7k dataset as input and output, respectively. We recorded the matching accuracy between the predicted theorem sequences and the annotated theorem sequences.

We define the matching degree as the ratio of the accurately predicted theorems in the predicted sequence to the number of theorems required for the problem. For instance, if a problem's theorem sequence includes 10 theorems and the predicted theorem sequence contains 8 of them, we consider the sequence matching degree to be 80\%. And ultimately, we obtain the following experimental results:

\begin{table}[h]
  \centering
  \begin{tabular}{c|c|c}
    \hline
    MODEL & AVERAGE(\%) & COMPLETE(\%)  \\
    \hline
    mT5 & 33.73 & 17.51 \\
     FLAN-T5-base & 59.37 & 33.59 \\
    FLAN-T5-large & 74.61 & 55.22 \\
    BART-base & 86.29 & 70.77 \\
    BART-large & 84.16 & 67.33 \\

    \hline
  \end{tabular}
  \caption{Prediction Matching Degree}
  \label{tab:PMD}
\end{table}

Where "average" represents the average matching degree of predicted theorem sequences for all questions. When the matching degree is 100\% (i.e., the predicted theorem sequence contains all the theorems required for that question), we consider the question solved with complete theorem sequence prediction, denoted as prediction complete. "complete" in the table signifies the percentage of questions with a predicted matching degree of 100\% in the FormalGeo7k dataset.

BART\cite{lewis2019bart} and T5\cite{raffel2020exploring} are both excellent Transformer architecture models, with BART being particularly suitable for sequence generation models, and T5 having greater versatility due to its "Text-to-Text" approach. According to the experimental results, we observed that, for the FormalGeo7k dataset, the BART-base pretrained model performs the best. Of course, both BART-large and T5-large also exhibit good predictive performance. Since our ultimate goal is to integrate the predicted theorem sequences into FGPS to achieve the best problem-solving results, we choose to use the BART-base pretrained model in the solver's solving process. This aims to explore the true problem-solving rate for the FormalGeo7k dataset.

\subsection{Experimental Results of FGeo-TP}

We used multiple processes to run the search algorithm, setting the maximum search depth to 15, beam size to 20, and a timeout of 600 seconds for each problem. For the solving methods and strategies employed by FGeo-TP, 
we employed both FW and BW methods, as well as BFS, DFS, RS, and BS search strategies. Table ~\ref{tab:FGeo-TP} presents the solving rates, unsolved rates, and timeout rates of these methods on the FormalGeo7k dataset.

\begin{table}[h]
  \centering
  
  \begin{tabular}{c|c|c|c|c}
    \hline
    method & strategy & solved & unsolved & timeout \\
    \hline
    FW & BFS & 68.16 & 1.96 & 29.88 \\
    FW & DFS & 67.36 & 1.87 & 30.76 \\
    FW & RS & 68.76 & 2.01 & 29.23 \\
    FW & BS & 60.06 & 4.40 & 35.54 \\
    BW & BFS & 80.12 & 1.81 & 18.07 \\
    BW & DFS & 79.55 & 2.14 & 18.31 \\
    BW & RS & \textbf{80.86} & \textbf{1.78} & \textbf{17.36} \\
    BW & BS & 79.06 &2.23 & 18.71 \\
    
    \hline
  \end{tabular}
  \caption{ the search results(\%) for FGeo-TP}
  \label{tab:FGeo-TP}
\end{table}

The experimental results reveal that FGeo-TP  achieves a maximum solving rate of 80.86\%, doubling the average solving rate compared to FGPS. Additionally, the maximum unsolved rate is only 4.4\%. Based on these results, we can confidently assert that FGeo-TP exhibits significant potential for solving geometric problems.

From the table, it is evident that FGeo-TP generally exhibits a higher solved rate in the backward search method compared to the forward search. However, in the case of FGPS, the solving results show the opposite situation. This discrepancy arises because, in the backward search method, each update of the final goal state is accompanied by updates to multiple child node states, consuming considerable time. When FGeo-TP provides new reasoning conditions, it significantly reduces the backward search process, thereby decreasing the timeout rate and improving the solving rate. For the forward search, the expansion of the initial condition set leads to an expansion of theorem selection paths, resulting in less optimization in areas with higher time complexity.

At the same time, we observe that the solving rate of FGeo-TP is lower than the matching degree of the theorem predictor for theorem sequences. Upon comparing the predicted theorem sequences with the actual solving theorem sequences, we find differences in the theorem order and the quantity of theorem uses. In geometric problem-solving, the same theorems with different usage orders may lead to different outcomes in solving success. Similarly, a question may require the repeated use of a particular theorem, while the theorem predictor only predicts whether this theorem should be used, without considering the frequency of use. Therefore, there is still room for improvement in the theorem predictor, and FGeo-TP may achieve even better solving performance.

\subsection{Solving Time and Step}
\begin{figure*}[ht]
	\centering
	\begin{minipage}{0.49\linewidth}
		\centering
		\includegraphics[width=0.9\linewidth]{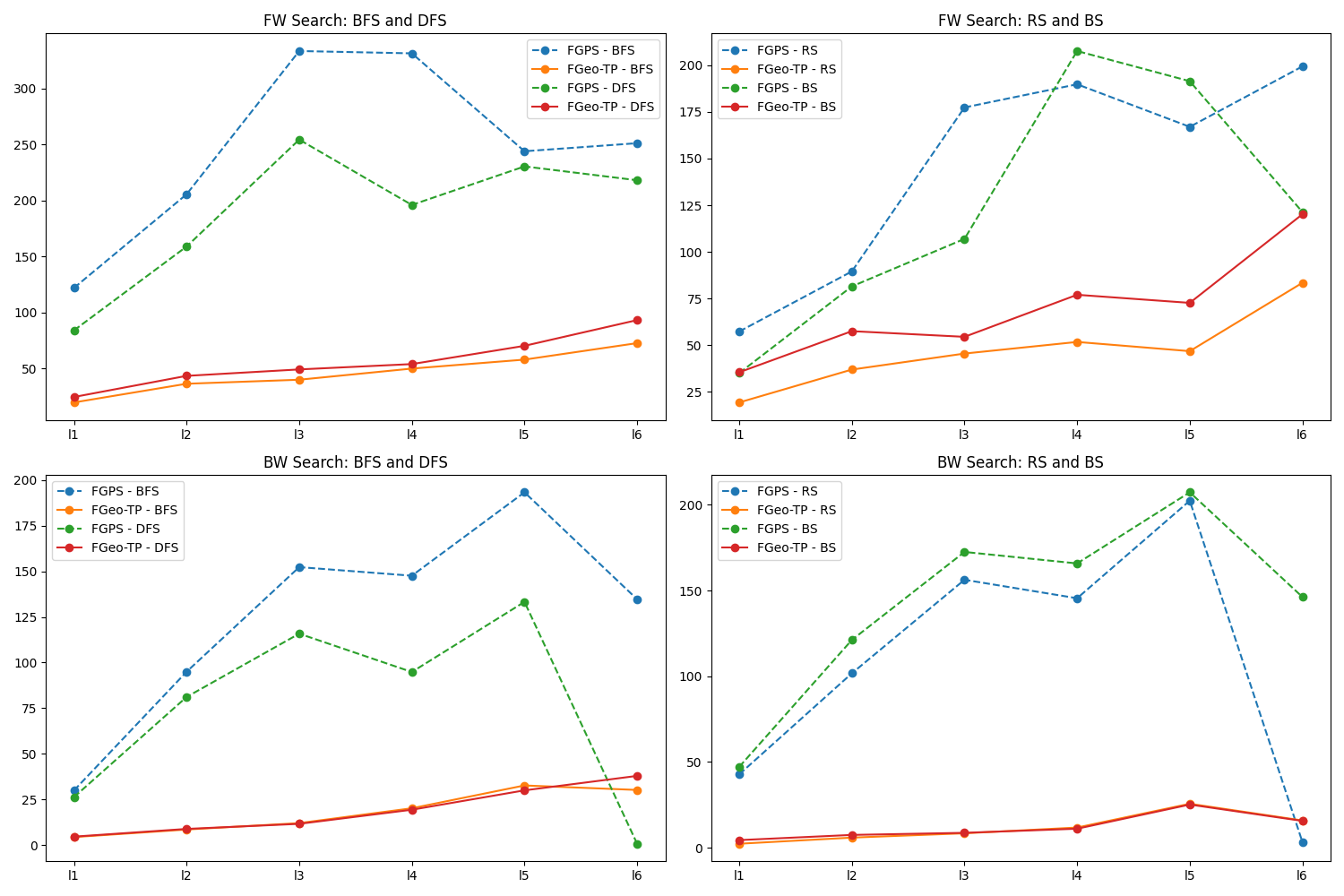}
		\caption{ solving time(s) at different difficulty levels.}
		\label{fig:fig3}
	\end{minipage}
	\begin{minipage}{0.49\linewidth}
		\centering
		\includegraphics[width=0.9\linewidth]{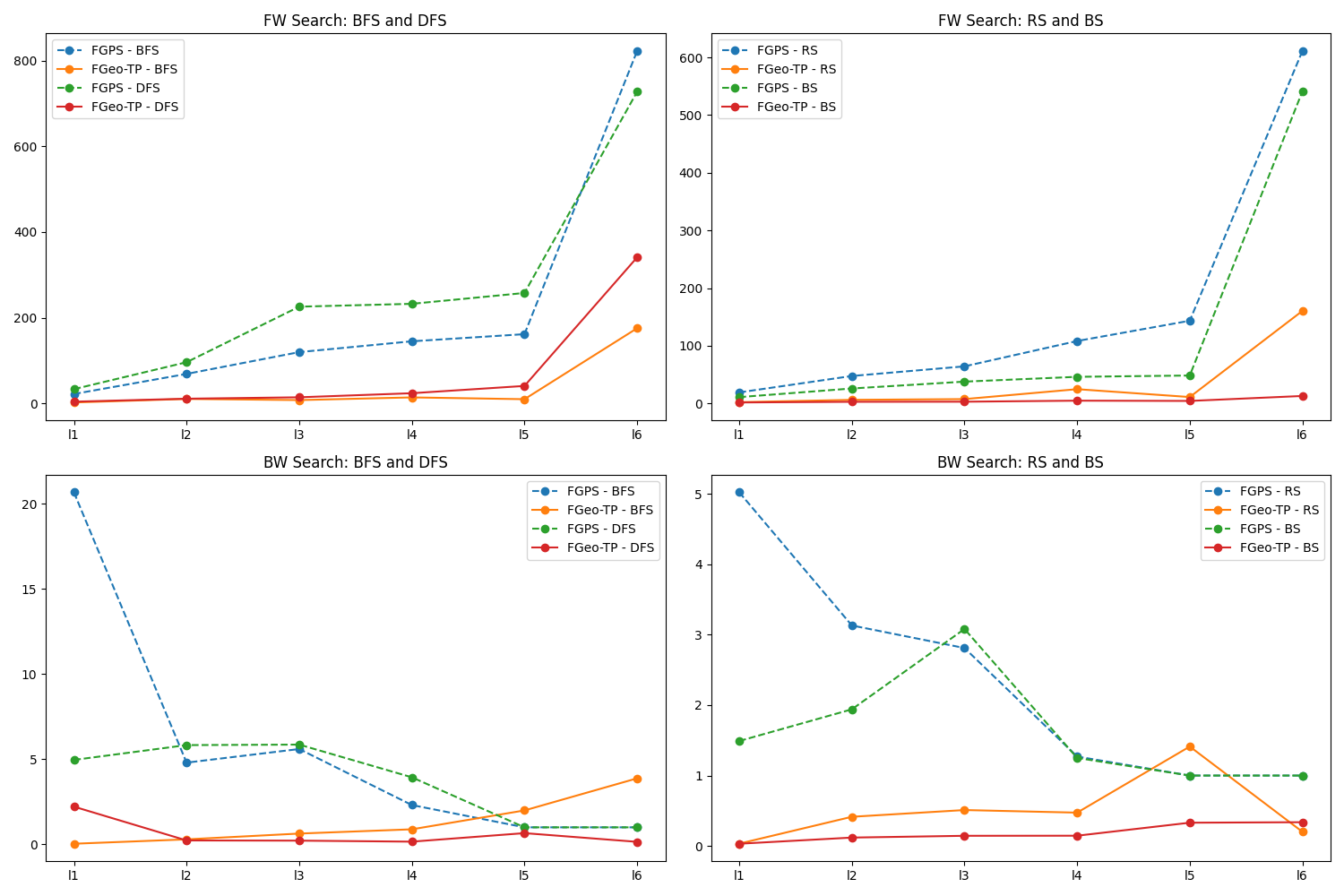}
		\caption{ solving step at different difficulty levels.}
		\label{fig:fig4}
	\end{minipage}
\end{figure*}

Based on the required length of theorem sequences for problem resolution, we have categorized the questions in FormalGeo7k into difficulty levels: $l_1$ ($\text{length} \leq 2$), $l_2$ ($3 \leq \text{length} \leq 4$), $l_3$ ($5 \leq \text{length} \leq 6$), $l_4$ ($7 \leq \text{length} \leq 8$), $l_5$ ($9 \leq \text{length} \leq 10$), and $l_6$ ($\text{length} \geq 11$). The corresponding quantities of questions for each difficulty level are 2407, 1898, 1247, 824, 313, and 292, respectively.
To investigate whether FGeo-TP optimizes the search space based on FGPS problem-solving, we compared the solving time and solution steps of the two for problems of varying difficulty levels. The experimental results on FormalGeo7k are illustrated in figure ~\ref{fig:fig3},figure ~\ref{fig:fig4},table~\ref{tab:time} and table~\ref{tab:step}.The detailed data can be found in the appendix \ref{appendix:date}.

\begin{table*}[ht] 
  \centering
   \begin{tabular}{c|c c c c |c c c c}
    \hline
      method & \multicolumn{4}{c}{FW} & \multicolumn{4}{c}{BW} \\
     \hline
    strategy & BFS & DFS & RS & BS  & BFS & DFS & RS & BS \\
    \hline
   FGPS & 185.05 & 132.35 & 92.21  & 58.76  & 57.67  & 46.45 & 65.82 &75.55  \\
FGeo-TP & 32.10 & 38.61 & 33.19 & 49.66 & 9.44 & 9.59 & 6.08 & 7.27\\
    
    \hline
  \end{tabular}
  \caption{ The average solving time results(s) for FGPS and FGeo-TP.}
  \label{tab:time}
\end{table*}

\begin{table*}[ht] 
  \centering
   \begin{tabular}{c|c c c c |c c c c}
    \hline
    method & \multicolumn{4}{c}{FW} & \multicolumn{4}{c}{BW} \\
     \hline
    strategy & BFS & DFS & RS & BS  & BFS & DFS & RS & BS \\
    \hline
   FGPS & 58.44 & 87.16  & 41.89   & 18.64   & 15.14   & 5.17  & 4.34  & 1.67   \\
FGEo-TP & 8.89 & 15.10 & 8.82 & 2.68 & 0.39 & 1.05 & 0.30 & 0.10\\
    
    \hline
  \end{tabular}
  \caption{ The average solving step results for FGPS and FGeo-TP.}
  \label{tab:step}
\end{table*}

From the graph, it is evident that, both in terms of solving time and solution steps, FGeo-TP exhibits superior performance compared to FGPS across the majority of difficulty levels for the given problems. Regarding the solving time, FGPS exhibits randomness due to its reliance on unstable searches. In contrast, FGeo-TP, with the integration of theorem prediction, bases its subsequent search space on given theorems. Consequently, the relationship between solving time and problem difficulty aligns more closely with human problem-solving behavior, reducing the overall average solving time by more than a quarter.

Under the guidance of theorem prediction, FGeo-TP rapidly identifies the direction for problem-solving, exhibiting significantly fewer solution steps than FGPS, especially in lower difficulty problems. However, as the complexity of the problems increases, the length of the necessary theorem sequences for solving these problems also extends, leading to a gradual weakening in the effectiveness of theorem prediction. This decline can be attributed to two main factors: Firstly, with the rise in problem difficulty, the search space for solutions grows exponentially, outpacing the rate at which theorem prediction can reduce this search space. Secondly, as the theorem sequences become longer, the accuracy of the theorem prediction decreases, consequently diminishing its utility in problem-solving.

\subsection{The Combined Approach of FGeo-TP and FGPS}
In the analysis of the problem-solving results using FGeo-TP, unexpected phenomena were observed. For certain problems, FGeo-TP failed to yield a solution, while the original FGPS successfully solved them. Consequently, additional experiments were conducted. Building upon FGeo-TP, a hybrid approach was employed, where, in case of timeouts, the search strategy was switched back to FGPS. The experimental outcomes, derived from the collaborative problem-solving efforts of FGeo-TP and FGPS, are presented in Table \ref{tab:FGPS and FGeo-TP}.

\begin{table*}[t] 
  \centering
   \begin{tabular}{c|c c c c |c c c c}
    \hline
    method & \multicolumn{4}{c}{FW} & \multicolumn{4}{c}{BW} \\
     \hline
    strategy & BFS & DFS & RS & BS  & BFS & DFS & RS & BS \\
    \hline
   FGeo-TP & 68.16 & 67.36 & 68.76 & 60.06 & 80.12 & 79.55& 80.86  &79.06 \\
FGeo-TP+FGPS & 71.56 & 72.14 & 73.48 & 65.07 & 81.09 & 80.67& 81.86&80.41\\
    \hline
  \end{tabular}
  \caption{ The average solving results(\%) for  FGeo-TP and the combined approach of FGeo-TP and FGPS.}
  \label{tab:FGPS and FGeo-TP}
\end{table*}

Based on the experimental results, it is observed that, for forward search , the combined approach of FGeo-TP and FGPS leads to an average improvement in problem-solving rate of approximately 3-5\% compared to FGeo-TP alone. Similarly, for backward search, the combined FGeo-TP and FGPS approach yields an average improvement in problem-solving rate of around 1\% compared to FGeo-TP alone. Further investigation was conducted to explore this anomalous phenomenon.

After comparing a considerable amount of experimental data, we identified the underlying reasons for the observed outcomes. The initial condition set of FGPS consists of the formal language annotations annotated in FormalGeo7k. These annotations include information provided by the problem statement, typically in limited quantity. Consequently, when FGPS selects a search path, it matches the available theorems based on the current condition set, placing feasible theorem sequences in a queue. Various search strategies will select different theorems from the queue for application. After applying a theorem, new conditions are obtained, and adding these conditions to the condition set generates a new condition set. The new set can once again match a batch of available theorem sequences, which are subsequently added to the previous theorem sequences.

Therefore, we observe that if FGeo-TP predicts the inclusion of non-essential theorems, the initial condition set may contain unnecessary conditions. This results in the generation of unnecessary condition sequences, diluting the probability of the solver selecting the correct theorem sequence. This significantly increases the prediction time, leading to timeouts. However, BFS is not affected by the addition of new theorems and continues to run the initially matched theorem sequences, ensuring that the desired theorems will be executed. From the experimental results of  the combined approach of FGeo-TP and FGPS, it is evident that only the FW method and BFS strategy shows a modest improvement of 3\%, while the other three strategies show an improvement of around 5\%. This observation indicates that BFS is less influenced by redundant theorems. 
In backward search, the effectiveness of  combined approach of FGeo-TP and FGPS has shown a modest improvement of 1\% compared to using FGeo-TP alone. The possible reason for this is the introduction of unnecessary conditions, leading to an increase in redundant algebraic equations during the solving process. Consequently, excessive and futile computations occur during equation solving, resulting in extended computation time and timeouts.

\section{Conclusion}

We combined a language model with FGPS to enhance the automatic problem-solving capability for plane geometry problems. Initially, we experimented with multiple pre-trained language models and selected BART-base as the optimal model due to its high theorem sequence prediction match rate of 86.29\%. Subsequently, we integrated the theorem predictor into the FGPS solving process, resulting in the development of FGeo-TP. Experimental results indicate that FGeo-TP achieves a problem-solving success rate of 80.86\% on the FormalGeo7k dataset, surpassing the solving rate of using FGPS alone by more than twice. Moreover, FGeo-TP's problem-solving process rapidly identifies the direction of solution, significantly reducing the number of problem-solving steps, and cutting the average problem-solving time by more than a quarter. In the future, we plan to further enhance the accuracy of the theorem predictor in predicting theorems, aiming to minimize unnecessary theorem predictions. Simultaneously, we will refine the dataset for IMO-level geometry problems and evaluate the performance of FGeo-TP on the FormalGeo-IMO dataset.

\sloppy
\section*{Acknowledgement}

This research was supported by NSFC Grant 12071282.

\appendix

\section{Experimental Results}

\begin{table*}[t]
\centering
\caption{Experimental Results}
\label{appendix:date}
\clearpage 
\begin{center}

\begin{tabular}{lccccccccccc}
\hline
 solver & metric & method & strategy & total & \( l_1 \) & \( l_2 \) & \( l_3 \) & \( l_4 \) & \( l_5 \) & \( l_6 \) \\
\hline
\multirow{16}{*}{\centering FGPS}  
  &\multirow{8}{*}{ time(s)}  
        & FW & BFS  & 185.05 & 121.97 & 205.51 & 333.33 & 331.26 & 243.92 & 251.11 \\
       & & FW & DFS  & 132.35 & 84.11 & 158.91 & 254.29 & 196.02 & 230.39 & 218.05 \\
       & & FW & RS  & 92.21 & 57.38 & 89.55 & 177.34 & 189.74 & 166.96 & 199.31 \\
       & & FW & BS  & 58.76 & 35.11 & 81.52 & 106.92 & 207.55 & 191.40 & 121.36 \\
       & & BW & BFS  & 57.67 & 30.00 & 94.97 & 152.26 & 147.55 & 193.21 & 134.65 \\
       & & BW & DFS  & 46.45 & 26.34 & 81.17 & 115.83 & 94.85 & 133.23 & 0.85 \\
       & & BW & RS & 65.82 & 42.79 & 101.71 & 156.27 & 145.50 & 202.37 & 3.37 \\
      &  & BW & BS  & 75.55 & 47.16 & 121.17 & 172.50 & 165.80 & 207.28 & 146.26 \\
    \cline{2-11}
   &\multirow{8}{*}{ step}  
       & FW & BFS & 58.44 & 21.85 & 68.67 & 119.56 & 144.80 & 161.44 & 822.18 \\
     &   &FW & DFS & 87.16 & 33.15 & 95.95 & 225.57 & 232.25 & 257.59 & 727.17 \\
     &  & FW & RS & 41.89 & 19.14 & 47.63 & 64.32 & 108.26 & 143.41 & 611.00 \\
     &  & FW & BS & 18.64 & 10.99 & 25.97 & 37.86 & 46.19 & 48.44 & 541.00 \\
     & &  BW & BFS & 15.14 & 20.68 & 4.80 & 5.60 & 2.31 & 1.00 & 1.00 \\
     & &  BW & DFS & 5.17 & 4.96 & 5.83 & 5.86 & 3.94 & 1.00 & 1.00 \\
    &  &  BW & RS & 4.34 & 5.02 & 3.13 & 2.81 & 1.27 & 1.00 & 1.00 \\
    &  &  BW & BS & 1.67 & 1.49 & 1.94 & 3.08 & 1.25 & 1.00 & 1.00 \\
\hline
\multirow{16}{*}{\centering FGeo-TP}  
  &\multirow{8}{*}{ time(s)}  
        & FW & BFS  & 32.10 & 19.83 & 36.51 & 40.11 & 50.05 & 58.05 & 72.76 \\
       & & FW & DFS  & 38.62 & 24.7 & 43.55 & 49.30 & 54.06 & 70.34 & 93.3 \\
       & & FW & RS  & 33.19 & 19.39 & 37.00 & 45.57 & 51.79 & 46.86 & 83.45 \\
       & & FW & BS  & 49.67 & 35.55 & 57.56& 54.53 & 77.04 & 72.71 & 120.13\\
       & & BW & BFS  & 9.44 & 4.99 & 8.63 & 12.14 & 20.27 & 32.75 & 30.29 \\
       & & BW & DFS  & 9.59 & 4.74 & 8.96 & 11.74 & 19.46 & 30.07 &37.99 \\
       & & BW & RS & 6.08 & 2.34 & 5.91 & 8.43 & 11.78 & 25.65 & 15.91 \\
      &  & BW & BS & 7.28 & 4.46 & 7.48 & 8.71 & 11.1 & 25.18 & 15.65  \\
    \cline{2-11}
   &\multirow{8}{*}{ step}  
       & FW & BFS & 8.897 & 2.07 & 10.27 & 7.61 & 13.87 & 9.68 & 175.242  \\
     &   &FW & DFS & 15.10 & 3.88 & 10.81 & 14.12 & 23.63 & 40.62 & 340.67  \\
     &  & FW & RS & 8.83 & 2.23 & 6.25 & 7.55 & 24.76 & 11.13 & 160.41 \\
     &  & FW & BS & 2.69 & 1.74 & 2.87& 3.04 & 4.84 & 4.39 & 12.978 \\
     & &  BW & BFS & 0.39& 0.04 & 0.29 & 0.63 & 0.88 & 1.99 & 3.88 \\
     & &  BW & DFS & 1.05 & 2.21 & 0.23 & 0.22 & 0.16 & 0.66 & 0.15  \\
    &  &  BW & RS &0.30 & 0.03 & 0.41 & 0.51 & 0.47 & 1.41 & 0.20 \\
    &  &  BW & BS & 0.10 & 0.03 & 0.12 & 0.14 & 0.15 & 0.33 & 0.34 \\
\hline
\end{tabular}
\end{center}
\footnotesize\textsuperscript{*} The experimental data for FGPS is derived from the FormalGeo\cite{Zhang2023FormalGeo}.
\end{table*}


\begin{thebibliography}{99}
\bibitem{Zhang2023FormalGeo}
Zhang X, Zhu N, He Y, et al. FormalGeo: The First Step Toward Human-like IMO-level Geometric Automated Reasoning[J]. arXiv preprint arXiv:2310.18021, 2023.

\sloppy
\bibitem{yang2023leandojo}
Yang K, Swope A M, Gu A, et al. Leandojo: Theorem proving with retrieval-augmented language models[J]. arXiv preprint arXiv:2306.15626, 2023.

\bibitem{petersen2021deep}
Brenden K. Petersen, et al. "Deep Symbolic Regression: Recovering Mathematical Expressions From Data Via Risk-Seeking Policy Gradients." ICLR 2021.

\bibitem{mundhenk2021symbolic}
Mundhenk T N, Landajuela M, Glatt R, et al. Symbolic regression via neural-guided genetic programming population seeding[J]. arXiv preprint arXiv:2111.00053, 2021.

\bibitem{polu2022formal}
Polu S, Han J M, Zheng K, et al. Formal mathematics statement curriculum learning[J]. arXiv preprint arXiv:2202.01344, 2022.

\bibitem{drori2022neural}
Drori I, Zhang S, Shuttleworth R, et al. A neural network solves, explains, and generates university math problems by program synthesis and few-shot learning at human level[J]. Proceedings of the National Academy of Sciences, 2022, 119(32): e2123433119.
\bibitem{Chen2021UniGeo}
Chen J, Li T, Qin J, et al. UniGeo: Unifying Geometry Logical Reasoning via Reformulating Mathematical Expression[J]. arXiv preprint arXiv:2212.02746, 2022.

\bibitem{lu2021intergps}
Lu P, Gong R, Jiang S, et al. Inter-GPS: Interpretable geometry problem solving with formal language and symbolic reasoning[J]. arXiv preprint arXiv:2105.04165, 2021.

\bibitem{zhang1995automated}
Zhang J Z, Chou S C, Gao X S. Automated production of traditional proofs for theorems in Euclidean geometry I. The Hilbert intersection point theorems[J]. Annals of Mathematics and Artificial Intelligence, 1995, 13(1-2): 109-137.

\bibitem{chen2021geoqa}
Chen J, Tang J, Qin J, et al. GeoQA: A geometric question answering benchmark towards multimodal numerical reasoning[J]. arXiv preprint arXiv:2105.14517, 2021.

\bibitem{polu2020generative}
Polu S, Sutskever I. Generative language modeling for automated theorem proving[J]. arXiv preprint arXiv:2009.03393, 2020.

\bibitem{seo2015solving}
Seo M, Hajishirzi H, Farhadi A, et al. Solving geometry problems: Combining text and diagram interpretation[C]//Proceedings of the 2015 conference on empirical methods in natural language processing. 2015: 1466-1476.

\bibitem{zhong2015interactive}
Zhong X, Fu H, Yu Y, et al. Interactive learning environment based on knowledge network of geometry problems[C]//2015 10th International Conference on Computer Science \& Education (ICCSE). IEEE, 2015: 53-58.

\bibitem{cao2022augmented}
Cao J, Xiao J. An augmented benchmark dataset for geometric question answering through dual parallel text encoding[C]//Proceedings of the 29th International Conference on Computational Linguistics. 2022: 1511-1520.

\bibitem{sachan2017learning}
Sachan M, Xing E. Learning to solve geometry problems from natural language demonstrations in textbooks[C]//Proceedings of the 6th Joint Conference on Lexical and Computational Semantics (* SEM 2017). 2017: 251-261.

\bibitem{wu2020knowledge}
Wu Q, Zhang Q, Fu J, et al. A knowledge-aware sequence-to-tree network for math word problem solving[C]//Proceedings of the 2020 conference on empirical methods in natural language processing (EMNLP). 2020: 7137-7146.

\bibitem{alvin2014synthesis1}
Alvin C, Gulwani S, Majumdar R, et al. Synthesis of geometry proof problems[C]//Proceedings of the AAAI Conference on Artificial Intelligence. 2014, 28(1).

\bibitem{alvin2017synthesis2}
Alvin C, Gulwani S, Majumdar R, et al. Synthesis of solutions for shaded area geometry problems[C]//The Thirtieth International Flairs Conference. 2017.

\bibitem{vaswani2017attention}
Vaswani A, Shazeer N, Parmar N, et al. Attention is all you need[J]. Advances in neural information processing systems, 2017, 30.

\bibitem{sun2020colake}
Sun T, Shao Y, Qiu X, et al. Colake: Contextualized language and knowledge embedding[J]. arXiv preprint arXiv:2010.00309, 2020.

\bibitem{liang2021mwpbert}
Liang Z, Zhang J, Wang L, et al. MWP-BERT: Numeracy-augmented pre-training for math word problem solving[J]. arXiv preprint arXiv:2107.13435, 2021.

\bibitem{wen1978decision}
Wen-Tsün W. On the decision problem and the mechanization of theorem proving in elementary geometry[J]. Scientia Sinica, 1978, 21(2): 159-172.

\bibitem{sutskever2014sequence}
Sutskever I, Vinyals O, Le Q V. Sequence to sequence learning with neural networks[J]. Advances in neural information processing systems, 2014, 27.

\bibitem{nevins1975plane}
Nevins A J. Plane geometry theorem proving using forward chaining[J]. Artificial Intelligence, 1975, 6(1): 1-23.

\bibitem{fawzi2022discovering}
Fawzi A, Balog M, Huang A, et al. Discovering faster matrix multiplication algorithms with reinforcement learning[J]. Nature, 2022, 610(7930): 47-53.

\bibitem{hao2022pgdp5k}
Hao Y, Zhang M, Yin F, et al. PGDP5K: A Diagram Parsing Dataset for Plane Geometry Problems[C]//2022 26th International Conference on Pattern Recognition (ICPR). IEEE, 2022: 1763-1769.

\bibitem{gelernter1963realization}
Gelernter H. Realization of a geometry-theorem proving machine[J]. Computers and thought, 1963: 134-152.

\bibitem{lewis2019bart}
Lewis M, Liu Y, Goyal N, et al. Bart: Denoising sequence-to-sequence pre-training for natural language generation, translation, and comprehension[J]. arXiv preprint arXiv:1910.13461, 2019.

\bibitem{raffel2020exploring}
Raffel C, Shazeer N, Roberts A, et al. Exploring the limits of transfer learning with a unified text-to-text transformer[J]. The Journal of Machine Learning Research, 2020, 21(1): 5485-5551.
\end{thebibliography}
\end{document}